\documentclass{article}

\usepackage[utf8]{inputenc} 
\usepackage[T1]{fontenc}    
\usepackage{hyperref}       
\usepackage{url}            
\usepackage{booktabs}       
\usepackage{amsfonts}       
\usepackage{nicefrac}       
\usepackage{microtype}      
\usepackage{amsmath}
\usepackage{amsfonts}
\usepackage{subfig}
\usepackage{graphicx}
\usepackage{adjustbox}
\usepackage{microtype}
\usepackage[inline]{enumitem}
\usepackage{algorithm}
\usepackage[noend]{algpseudocode}
\usepackage{dsfont}

\newtheorem{proposition}{Proposition}

\newcommand\independent{\protect\mathpalette{\protect\independenT}{\perp}}
\def\independenT#1#2{\mathrel{\rlap{$#1#2$}\mkern2mu{#1#2}}}
\def\+#1{\mathbf{#1}}
\usepackage{xcolor}

\title{Time Adaptive Gaussian Model}

\author{Federico Ciech$^1,*$ and  Veronica Tozzo$^{2,3,*}$}
\date{\small
$^1$ Department of Bioengineering, Robotics, Informatics and System Engineering, Universit\`a degli Studi di Genova\\
$^2$ Center for System Biology and Department of Pathology, Massachusetts General Hospital\\
$^3$ Department of Systems Biology, Harvard Medical School\\
$^*$ These authors equally contributed to this paper.
}

\begin{document}
\maketitle

\begin{abstract}
  Multivariate time series analysis is becoming an integral part of data analysis pipelines. Understanding the individual time point connections between covariates as well as how these connections change in time is non-trivial. To this aim, we propose a novel method that leverages on Hidden Markov Models and Gaussian Graphical Models --- Time Adaptive Gaussian Model (TAGM). Our model is a generalization of state-of-the-art methods for the inference of temporal graphical models, its formulation leverages on both aspects of these models providing better results than current methods. In particular,it  performs pattern recognition by clustering data points in time; and, it finds probabilistic (and possibly causal) relationships among the observed variables. Compared to current methods for temporal network inference, it reduces the basic assumptions while still showing good inference performances. 
\end{abstract}

\section{Introduction}
%
%
The inference of temporal networks has started to become a common topic in the last few years \cite{guo2011joint,danaher2014joint, hallac2015network, hallac2017network, hallac2017toeplitz, chang2019graphical, tomasi2019temporal, tomasi2018latent}. 
Current method approach the problem by taking a multi-variate time series and dividing it in chunks \cite{foti2016sparse,hallac2017network,tomasi2018latent}. 
Each chunk is assumed to be a short enough period of time that all its points are identically and independently sampled from a unique distribution. Such distribution, when the variables are continuous, can be represented as a Gaussian Graphical Model (GGM), where the conditional independency patterns of variables are encoded as edges of graphs that evolve in time.
These approaches show good performances, but the assumption that the time points in each chunk are i.i.d. is most often is not true. 

In this paper, we propose Time Adaptive Gaussian Model (TAGM), a combination of GGMs with Hidden Markov Models (HHMs) \cite{baum1966statistical} that allows to easily consider the sequence of single time points and relax the chunk assumption. It also allows us to obtain clusters of time points as well as repeated evolving patterns of graphs that may be impossible to obtain with current state-of-the-art methods.

\begin{figure}[t]
  \includegraphics[width=1\textwidth]{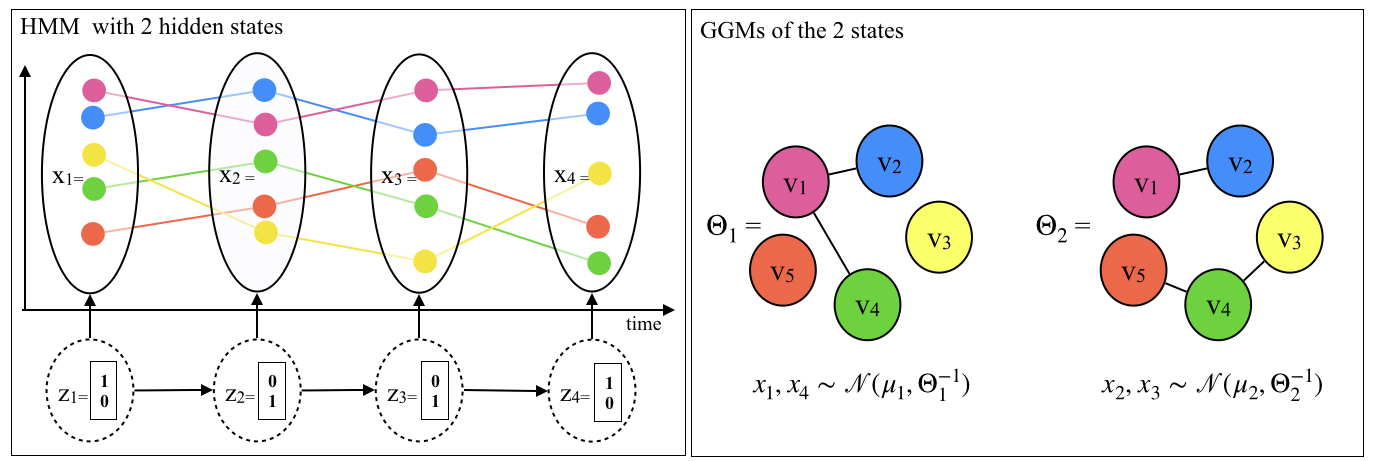}
  \caption{TAGM schematic representation. To each temporal observation is associated a state. Each state is characterized by an underlying distribution which is represented by a graphical model. }
  \label{fig:hmm}
  \end{figure}

A schematic representation of the model is presented in Figure~\ref{fig:hmm}. Here, we are considering an HMM at 2 states, which by looking at the left panel, presents as the sequence 1, 2, 2, 1. Given the latent states,  observations $x_1$ and $x_4$ belong to an underlying distribution while $x_2$ and $x_3$ belong to another one. Given the Markov chain that connects the latent state we can assume that all these observations are independent and, thus, use them to infer two GGMs that models the probability distribution of cluster 1 and cluster 2 (right panel Figure~\ref{fig:hmm}). The inferred GGMs provide us more information on how, within each latent state, the variables are dependent to each other. 

Our approach, given a multi-variate time-series, is able to 

\begin{enumerate}
    \item \textbf{cluster temporal data points considering sequentiality}. Note that, TAGM does not intend to group time series based on their morphology. Rather, it \emph{clusters} the observations within the time series based on their similarities and the order they appear. 
Such approach, in literature, is either formulated as a standard clustering problem on the time points or as a longitudinal clustering on the time-series \cite{macqueen1967some, everitt2014finite, ng2002spectral}. 

\item \textbf{Infer temporal-dependent conditional dependencies among variables}: TAGM relaxed the chunk assumption common to the inference methods available in literature \cite{hallac2017network, tomasi2018latent} thus inferring a time-varying network that adapts at each observation. Note that TAGM assumes a sequentiality of the states, while
in \cite{lotsi2013high,everitt2014finite} the authors  combined GGMs with Gaussian Mixture Models. This approach is more prone in clustering the points in classes which can be seen as an unsupervised extension of the Joint Graphical Lasso \cite{danaher2014joint}. 
\end{enumerate}

%
%

 %
%
We show on synthetic data that the model performs better than current state-of-the-art methods on both tasks (1) and (2). We argue that the general characteristics of TAGM make it a trustful model that may be applied in a variety of applicative domains.

\paragraph{Related work}
For time series clustering, to our knowledge, HMM \cite{baum1966statistical} is the only clustering method which considers also sequentiality, as state-of-the-art methods typically cluster points based on the feature differences \cite{macqueen1967some, everitt2014finite, ng2002spectral}.
The inference of time-varying network has being tackled recently in literature \cite{foti2016sparse,hallac2017network,tomasi2018latent} by dividing the time-series in chunks with the strong assumption that all the time points within a chunk are i.i.d. TAGM overcomes this assumption proposing a more elegant way of inferring evolving networks as well as their pattern of evolution. By inferring $K$ different graphs we reach a deeper level of understanding on the states that allows us to gain insights on the system under analysis \cite{hallac2017toeplitz}. Lastly, 
state-of-the-art prediction methods on time-series \cite{sims1980macroeconomics, hochreiter1997long, 8569801, vert2004primer} commonly assume the relations among the past values of the variables and the present values of each variable to be constant in time.
An idea similar to TAGM was proposed with Gaussian Mixture Models (GMMs) \cite{everitt2014finite} where they combined GMM with GGMs \cite{lotsi2013high}. The use of GMMs though would not allow to explicitly consider sequentiality and it is therefore not suited for the analysis of time-series.  In literature, we found two examples that explicitly consider non-stationarity and sequentiality in a setting similar to ours.
\section{Preliminaries}
Consider a complex non-stationary system where, with
non-stationarity, we intend a change in time of the underlying distribution of observations. Of this system we observe $N$ temporal observations, each of these observation $n=1, \dots, N$ is a vector $\+x_n \in \mathbb{R}^d$ of $d$ variables sampled from an unknown distribution.
In the following we will denote vectors with bold letters $\+x$, matrices with capital letters $X$ and sequences of vectors or matrices with bold capital letters $\+X$.


\subsection{Hidden Markov Model}
HMMs are statistical models widely applied sequential data structures. HMMs assume that the series of observations is generated by a given number $K$ of (hidden) internal states
which follow a Markov process (see left panel of Figure~\ref{fig:hmm}) \cite{baum1966statistical}.
Consider the $N$ sequential (temporal) observations, 
we pair each of them with a hidden (latent) state $\mathbf{z}_n = \sum_{i=1}^K\mathds{1}_{\{i=k\}} \+e_i$, where $\+e_i$ is the $K$-dimensional natural basis which has a non-zero component only at position $i$ and $k$ is the \emph{cluster} label of observation $n$. We use the notation $z_{n,k}$ to indicate the $k$-th positional value of the vector $\+z_{n}$. 
Through their hidden states, the observations $\mathbf{x}_n$ and $\mathbf{x}_{n+1}$ become independent given their states (see Figure~\ref{fig:hmm} left panel). The hidden states, on the other hand, follow a Markov chain process which satisfies the conditional independence property
$
    \+z_{n+1}\independent \+z_{n-1}|\+z_n
$. \\
The HMM joint distribution on both the observations $\mathbf{X} = (\mathbf{x}_1, \mathbf{x}_2, \dots, \mathbf{x}_N)$   and the latent variables of $\mathbf{Z} = (\mathbf{z}_1, \mathbf{z}_2, \dots, \mathbf{z}_N)$ is then given by
\begin{equation}\label{eq:probability}
    p(\+X,\+Z |\+\pi,A,\+\phi)=
    p(\+z_1 |\pi)\Big[\prod_{n=2}^Np(\+z_n|\+z_{n-1}, A)\Big]\prod_{n=1}^Np(\+x_n|\+z_n, \phi).
\end{equation}
The probability $p(\+z_1|\+\pi)=\prod_{k=1}^K\pi_k^{z_{1,k}} \quad\text{with}\quad\sum_k\pi_k=1$ is the initial latent node $\+z_1$ probability, which differs from the other states as there is no parent node. Thus, its marginal distribution is embodied by a vector of probabilities $\+\pi$
whose elements $\pi_k \equiv p(z_{1,k} = 1)$ represents the probability of the first observation to belong to the $k$-th state.\\
%
The probability $p(\+z_n|\+z_{n-1,A})=\prod_{k=1}^K\prod_{j=1}^KA_{j,k}^{z_{n-1,j}z_{n,k}}$ is the \textit{transition probability} of moving from one state to the other where $A \in [0,1]^{K \times K}$ is the \textit{transition matrix} that we assume to be constant in time.It is defined as $
A_{j,k} = p(z_{n,k} = 1|z_{n-1,j}=1)$ with $0\leq A_{j,k}\leq 1$ and $ \sum_k A_{j,k}=1$. \\
 Lastly, $p(\+x_n|\+z_n,\+\phi) = \prod_{k=1}^Kp(\+x_n|\phi_k)^{z_{n,k}}$, are the \textit{emission probabilities} where $\+\phi=\{\phi_1,\dots,\phi_K\}$ is a set of $K$ different parameters governing the distributions, one for each of the possible $K$ states. 

\subsection{Gaussian Graphical Models}
%
%
GGMs are typically employed in the analysis of multivariate problems where one seeks to understand the relationships among variables. A GGM is a probability distribution which factorizes according to an undirected graph
whose set of edges univocally determines a multivariate normal distribution $\mathcal{N}(\mu, \Sigma)$. Indeed, the \emph{precision matrix}, $\Theta = \Sigma^{-1}$ encodes the conditional independence between pairs of variables, and $\Theta(i,j) =0$ implies the absence of an edge in the graph. 
Thus, $\Theta$ is the weighted adjacency matrix of the graph (see right panel Figure~\ref{fig:hmm}) \cite{lauritzen1996graphical}.\\
%
Given $M$ observations of $d$ variables  \(\mathbf{X} \in \mathbb{R}^{M\times d}\) we aim at inferring the underlying graph corresponding to the precision matrix $\Theta$. 
In order to perform such inference we need to assume that the underlying graph is sparse. This is due to the combinatorial nature of the problem that requires to be constrained to have identifiability guarantees \cite{friedman2008sparse}. 
The most common way of inferring such graph is through the \emph{Graphical Lasso} (GL) \cite{friedman2008sparse}, a penalized Maximum Likelihood Estimation (MLE) method that solves the following problem 
\begin{equation}\label{GLexpr}
\underset{\Theta \succ 0}{\text{argmin}} ~~\text{tr}(\Theta S) -  \text{log det}(\Theta) + \lambda\|\Theta\|_{1, od}
\end{equation}
where $S$ is the empirical covariance matrix of the input data defined as $S = \frac1n X^\top X$, $\text{tr}(\Theta S)~-~\text{log det}(\Theta) $ is the negative log likelihood of the multivariate normal distribution and $\|\cdot\|_{1, dot}$ is the off-diagonal $\ell_1$-norm that imposes sparsity on the precision matrix $\Theta$ without considering the diagonal elements.

\section{Time Adaptive Gaussian Model}

TAGM presents an elegant way of inferring evolving networks as well as their pattern of evolution. Consider the system presented in previous section, we have $N$ temporal observations that are connected through $K$ latent states. Each of this state is connected to an unknown distribution that we want to model as a graphical model. 
 By inferring $K$ different graphs we reach a deeper level of understanding on the intra-dependency patterns of the states that allows us to gain insights on the system under analysis \cite{hallac2017toeplitz}. 

TAGM is a combination of HHMs with GGMs, which can be straightforwardly derived by assuming that 
%
 all the observations belonging to a cluster $k$ are i.i.d. and multivariate normally distributed.
 Thus, we assume that the emission probabilities are
 \begin{equation*}
    p(\+x_n|\+z_n,\+\phi) = \prod_{k=1}^K  \mathcal{N}(\+x_n| \mu_k, \Theta_k^{-1})^{z_{n,k}}
 \end{equation*}
 in such a way to have an explicit correspondence between the distribution of each state of the HMM and a graph, modeled through the precision matrix $\Theta$. Thus, the parameters $\phi$ are equal to the sequences $\mu= (\mu_1, \dots, \mu_K)$ and $\+\Theta= (\Theta_1, \dots, \Theta_K)$.\\
In order to obtain the Graphical Lasso form we add a sparsity constraint by multiplying the joint probability distribution in Equation~\eqref{eq:probability} with a Laplacian prior on the precision matrix. The Laplacian prior would also provide more interpretable results as we keep only the strongest edges of the final graphs.  The posterior is defined as
\begin{align*}
   p(\+\pi,A,\+\mu, \+\Theta|\+X,\+Z)\propto &p(\+z_1 |\pi)\Big[\prod_{n=2}^Np(\+z_n|\+z_{n-1}, A)\Big]\\
   &\prod_{n=1}^N\prod_{k=1}^K\mathcal{N}(\+x_n| \mu_k, \Theta_k^{-1})^{z_{n,k}}e^{-\frac{\lambda}{2}||\Theta_k||_{1,od}}.
\end{align*}
where with the notation $\+z_n$ and $\+z_{n-1}$ we hid a product on $z_{n,k}$ and $z_{n-1,j}$ for $k,j=1, \dots, K$. 
%
%
In order to fit such distribution on data we need to detect the best set of parameters $\theta= \{ \pi, A,\mathbf{\mu},\+\Theta\}$ through a Maximum A Posteriori approach. 
To marginalize over the latent variables $p(\+X|\+\theta)=\sum_{\+Z}p(\+X,\+Z|\+\theta)$ 
we employ the \emph{expectation maximization} (EM) algorithm,  in its particular version known as Baum's algorithm \cite{baum1970maximization}. 
The \emph{expectation maximization} (EM) algorithm \cite{baum1970maximization} is an iterative algorithm which alternates two steps: the \emph{E-step} and the \emph{M-step}. It starts with some initial selection for the model parameters, which we denote by $\theta_{\text{old}}$. Then from the posterior distribution over the latent variables $p(\+X,\+Z|\+\theta_{\text{old}})$ it evaluates the expectation of the complete-data log-likelihood as a function of the parameters $\theta= \{\+\pi,A,\+\mu,\+\Theta\}$:
\begin{align*}
  Q(\theta,\theta_{\text{old}})=&\gamma(\+z_n)\ln(\+\pi)+\sum_{n=2}^N   \xi(\+z_{n-1},\+z_n)\ln A+\\
  &\sum_{n=1}^N \sum_{k=1}^K \left[\gamma(z_{n,k})\ln \mathcal{N}(\+x_n| \mu_k, \Theta_k^{-1})-\frac{1}{2}\lambda ||\Theta_k||_{1,od} \right]
\end{align*}
where $\gamma(z_{n,k})$ and $\xi(z_{n-1,j},z_{n,k})$ come from the definition
\begin{align*}
    \gamma(\+z_n)&= p(\+z_n|\+X,\+\theta_{\text{old}})\\
    \xi(\+z_{n-1},\+z_n)&= p(\+z_{n-1},\+z_n|\+X,\+\theta_{\text{old}}).
\end{align*}
Denoting $\gamma(z_{n,k})$ the conditional probability of $z_{n,k}= 1$, with a similar use of notation for $\xi(z_{n-1,j},z_{n,k})$. Because the expectation of a binary random variable is just the probability that it takes the value 1, we have
\begin{align*}
    \gamma(z_{n,k})&=\mathbb{E}[z_{n,k}]=\sum_{\+z}\gamma(\+z)z_{n,k}\\
    \xi(z_{n-1,j},z_{n,k})&= \mathbb{E}[z_{n-1,j}z_{n,k}]= \sum_{\+z}\gamma(\+z)z_{n-1,k}z_{n,k}.
\end{align*}
The goal of the \emph{E-step} is to evaluate the quantities $\gamma(\+z_n)$ and $\xi(\+z_{n-1},\+z_n)$. While the \emph{M step} maximizes $Q(\theta,\theta_{\text{old}})$ with respect to the parameters $\+\theta$ in which we treat $\gamma(\+z_n)$ and $\xi(\+z_{n-1},\+z_n)$ as constants. 

\subsection{E step} An efficient algorithm to evaluate the quantities $\gamma(\+z_n)$ and $\xi(\+z_{n-1},\+z_n)$ is the \textit{forward-backward} algorithm described in \cite{baum1970maximization}. Here we want to emphasize the most relevant formulas which characterizes the algorithm. If the reader is interested in more details and how to derive them we refer to \cite{bishop2006pattern}.
 
Using Bayes’ theorem, we have
\begin{equation}\label{gammazn}
    \gamma(\+z_n)=\frac{p(\+X|\+z_n)p(\+z_n)}{p(\+X)}= \frac{p(\+x_1,\dots,\+x_n,\+z_n)p(\+x_{n+1},\dots,\+x_N|\+z_n)}{p(\+X)}= \frac{\alpha(\+z_n)\beta(\+z_n)}{p(\+X)}.
\end{equation}
where we have defined
\begin{align*}
    \alpha(\+z_n)&=p(\+x_1,\dots,\+x_n,\+z_n)\\
    \beta(\+z_n)&=p(\+x_{n+1},\dots,\+x_N|\+z_n).
\end{align*}
$\alpha(\+z_n)$ is also called \emph{forward process} while $\beta(\+z_n)$  \emph{backward process}. It is possible to prove that by the conditional independence assumption $\+z_{n+1}\independent \+z_{n-1}|\+z_n$ the forward and backward processes are characterized by two recursive equations:
\begin{align}
\alpha(\+z_n)&=p(x_n|\+z_n)\sum_{\+z_{n-1}}\alpha(\+z_{n-1})p(\+z_n|\+z_{n-1}),\label{alp}\\
\beta(\+z_n)&=\sum_{\+z_{n+1}}\beta(\+z_{n+1})p(\+x_{n+1}|\+z_{n+1})p(\+z_{n+1}|\+z_{n}),\label{bet}
\end{align}
with the initial conditions:
\begin{align*}
\alpha(\+z_1)&=\prod_{k=1}^K\{\pi_k\mathcal{N}(\+x_1|\mu_k,\Theta^{-1}_k)\}^{z_{1,k}}\\
\beta(\+z_N)&=(1,\dots,1).
\end{align*}
If we sum both sides of \eqref{gammazn} over $\+z_n$, and use the fact that the left-hand side is a normalized distribution, we obtain
\[
    p(\+X)=\sum_{\+z_n}\alpha(\+z_n)\beta(\+z_n)=\sum_{\+z_N}\alpha(\+z_N).
\]
Using similar arguments we have
\[
    \xi(\+z_{n-1},\+z_n)=\frac{\alpha(\+z_{n-1}) p(\+x_{n}|\+z_{n})p(\+z_n|\+z_{n-1})\beta(\+z_n)}{p(\+X)}.
\]

For moderate lengths of chain the calculation of $\alpha(\+z)$ can go to zero exponentially quickly. We therefore work with re-scaled versions of $\alpha(\+z)$ and $\beta(\+z)$ whose values remain of order unity. The corresponding scaling factors cancel out when we use there re-scaled quantities in the EM algorithm.  

We define a normalised version of $\alpha(\+z)$ as
\[
\hat{\alpha}(\+z_n) = p(\+z_n| \+x_1, \dots, \+x_n) = \frac{\alpha(\+z_n)}{p( \+x_1, \dots, \+x_n)}
\]
which we expect to be well behaved numerically because it is a probability distribution over $K$ variables for any value of $n$. In order to relate to the original $\alpha(\+z)$ variables we introduce scaling factors 
\[
c_n = p( \+x_n| \+x_1, \dots, \+x_{n-1})
\]
and therefore
\[
p( \+x_1, \dots, \+x_n) = \prod_{m=1}^nc_m.
\]
From the $\alpha$ and $\beta$ recursive equations \eqref{alp} and \eqref{bet} its scaled formula $\hat{\alpha}(\+z)$ becomes
\[
\hat{\alpha}(\+z_n) = \frac{p(\+x_n |\+z_n) \sum_{\+z_{n-1} }\hat{\alpha}(\+z_{n-1})p(\+z_n |\+z_{n-1})}{c_n}
\]
and, similarly 
\[
\hat{\beta}(\+z_n) = \frac{\sum_{\+z_{n+1} }\hat{\beta}(\+z_{n+1})p(\+x_{n+1} |\+z_{n+1}) p(\+z_{n+1} |\+z_{n})}{c_{n+1}}.
\]
Note that for the computation of $\beta$s we can recur to the scaling factors we computed in the $\alpha$ phase. 
We can also notice that the probability distribution of $\+X$ becomes 
\[
p(\+X) = \prod_{n=1}^N c_n
\]
and that 
\begin{align*}
\gamma(\+z_n) &= \hat{\alpha}(\+z_n)\hat{\beta}(\+z_n)\\
\xi(\+z_{n-1},\+z_n ) &= c^{-1}_n \hat{\alpha}(\+z_n)p(\+x_n |\+z_n)p(\+z_n|\+z_{n-1})\hat{\beta}(\+z_{n}).
\end{align*}


\subsection{M step:} Given $\gamma(\+z_n)$ and $\xi(\+z_{n-1},\+z_n)$ computed in the E step, the M step finds the optimal parameters $\+\theta$. To maximize respect to $\pi$ and $\+A$ we keep the addends of $Q(\theta,\theta_{\text{old}})$ which are directly interested in, where we use the appropriate Lagrangian to take into account of the constraints.
\begin{align*}
    \frac{\partial Q(\pi_k)}{\partial \pi_k}&= \frac{\partial \Big(\sum_{k=1}^K\gamma(z_{1,k})\ln(\pi_k)+ \lambda(1-\sum_{k=1}^K\pi_k)\Big)}{\partial \pi_k}= \frac{\gamma(z_{1,k})}{\pi_k}-\lambda=0
\end{align*}
We multiply both sides by $\pi_k$ and summing over all $k\in \{1,\dots,K\}$ and obtain $\lambda= \sum_{j=1}^K\gamma(z_{1,k})$ using that $\sum_{j=1}^K\pi_j=1$. Therefore 
\[
    \pi_k = \frac{\gamma(z_{1,k})}{ \sum_{j=1}^K\gamma(z_{1,k})}.
\]
Analogously for $\+A$
\begin{align*}
    \frac{\partial Q(A_{j,k})}{\partial A_{j,k}}&= \frac{\partial \Big(\sum_{n=2}^N\sum_{j=1}^K\sum_{k=1}^K \xi(z_{n-1,j},z_{n,k})\ln A_{j,k}+ \lambda(1-\sum_{k=1}^KA_{j,k})\Big)}{\partial A_{j,k}}\\& = \frac{\sum_{n=2}^N\xi(z_{n-1,j},z_{n,k})}{ A_{j,k}}-\lambda=0
\end{align*}
Similarly as before we multiply both sides by $A_{j,l}$ and summing over all $l\in \{1,\dots,K\}$ and obtain $\lambda= \sum_{l=1}^K\sum_{n=2}^N\xi(z_{n-1,j},z_{n,l})$ using that $\sum_{l=1}^K A_{j,l}=1$. Therefore 
\[
    A_{j,k} = \frac{\sum_{n=2}^N\xi(z_{n-1,j},z_{n,k})}{ \sum_{l=1}^K\sum_{n=2}^N\xi(z_{n-1,j},z_{n,l})}.
\]
When we differentiate by $\mu_k$ we have the same equation as in the Gaussian mixture model, therefore we give directly the result
\[
    \mu_k=\frac{\sum_{n=1}^N\gamma(z_{n,k})\+x_n}{\sum_{n=1}^N\gamma(z_{n,k})}.
\]

The maximization of $\Theta_k$ is performed in the main paper in Section 3.

For a detailed explanation of all the properties of this algorithm we refer to \cite{neath2013convergence}. The E step corresponds to evaluating the expected value of the log-likelihood at $\+\theta$.
Given the values $\theta_{\text{old}}$, \textit{i.e.}, the parameter values at the previous iteration,
the expectation is defined by the function 
\begin{align}\label{Q}
\begin{split}
    Q(\theta,\theta_{\text{old}})&=\gamma(\+z_n)\ln(\+\pi)+\sum_{n=2}^N   \xi(\+z_{n-1},\+z_n)\ln A \\
    &+\sum_{n=1}^N \sum_{k=1}^K \left[\gamma(z_{n,k})\ln \mathcal{N}(\+x_n| \mu_k, \Theta_k^{-1})-\frac{1}{2}\lambda ||\Theta_k||_{1,od} \right]
    \end{split}
\end{align}
%
In the functional, $\gamma(\+z_n)= p(\+z_n|\+X,\+\theta_{\text{old}})$ and $\xi(\+z_{n-1},\+z_n)= p(\+z_{n-1},\+z_n|\+X,\+\theta_{\text{old}})$ are the expectations on the latent variables and they are typically computed by a \emph{forward-backward} algorithm \cite{bishop2006pattern}.\\
%
%
%
Once the computation of the expectation is performed, the M step consists in  finding the $\+\theta$ values which maximize the function $Q$.
The derivation of the new parameters $\pi, A$ and $\mu$ is straightforward from literature \cite{bishop2006pattern}. The maximization of Equation~\eqref{Q} w.r.t. $\+\Theta$, instead requires further attention and it can be shown that, given the imposition of the Laplacian prior, reduces to the Graphical Lasso (Equation \eqref{GLexpr}) with few algebraic manipulations. 
Furthermore, the maximization of $\+\Theta$ can be performed separately for each $\Theta_k$ as, given $k$, all the maximizations are independent between each other. Thus, if we indicate as $\bar{\+x}_i=\+x_i-\+\mu_k$ the centered observations belonging to cluster $k$ 
the joint log-likelihood at fixed $k$ writes out as
\begin{align}
    \small
    &\sum_{n=1}^N\gamma(z_{n,k})\ln\mathcal{N}(\mathbf{x}_n|\mu_k,\Theta_k^{-1})-\frac{1}{2}\lambda || \Theta_k||_{1,od}\notag\\
    =& \sum_{n=1}^N\frac{\gamma(z_{n,k})}{2}\ln\text{det}\Theta_k-\frac{1}{2}\text{tr}\Big(\sum_{n=1}^N\gamma(z_{n,k})(\bar{\+x}_n\bar{\+x}_n')\Theta_k\Big)\notag-\frac{1}{2}\lambda || \Theta_k||_{1,od}\notag\\
    =& \sum_{n=1}^N\frac{\gamma(z_{n,k})}{2}\Big[\ln\text{det}\Theta_k-\text{tr}\Big(\Tilde{S}_k\Theta_k\Big)-\Tilde{\lambda}_k|| \Theta_k||_{1,od}\Big]\label{Qthk}
\end{align}
where $\Tilde{S}_k=\frac{\sum_{n=1}^N\gamma(z_{n,k})(\bar{\mathbf{x}}_n,\bar{\mathbf{x}}_n')}{\sum_{n=1}^N\gamma(z_{n,k})}$ is the weighted empirical covariance matrix, $\Tilde{\lambda}_k=\frac{\lambda}{\sum_{n=1}^N\gamma(z_{n,k})}$ and
\begin{equation}\label{sigmastarlasso}
 \Theta_k= \arg\max_{\Theta_k}\Big\{ \ln\text{det}\Theta_k-\text{tr}\Big(\Tilde{S}_k\Theta_k\Big)-\Tilde{\lambda}_k|| \Theta_k||_{1,od} \Big\}
\end{equation}
which is equivalent to a Graphical Lasso \cite{friedman2008sparse}.  Equation~\eqref{sigmastarlasso}  is a convex functional having guarantees of reaching a global optimum. Differently, Equation~\eqref{Q} is non-convex and depending on the initialization different local optima may be reached. 
\section{TAGM extensions}
The proposed approach can benefit from two types of extensions.  
The first one consists in higher-order Markov relationships among latent states. This can be achieved following \cite{hadar2009high}, with the main drawback of a much higher computational time for learning as the transition matrix and the corresponding initial state dimensions increase. 
The second extension consists in an online learning version that will allow to employ the model in a more applicative setting where we may deal with high-frequency data \cite{chis2015adapting}. Indeed, in a situation where new observations arrive at a high rate (every second or even millisecond) we want to be able to fine tune the model online in order to consider such observations instantaneously. This will allow to promptly gain insights on data and possibly predict the next time point. 
The weakness of this extension is that it requires an approximation 
which makes the updated parameters less accurate respect to the batch (original) version. 
We discuss in further details these two extensions in Sections 8 and 9  of the Supplementary material. In particular, we present two experiments where we compare them with TAGM that show the potential weaknesses.

\subsection{Higher order extension: Memory Time Adaptive Gaussian model (MemTAGM)}
Sometimes real world applications have events which rely on their past realizations. Therefore we can exploit more information from data if we consider a higher-order Markov process whose $\+z_n$ state probability does not depend only on $\+z_{n-1}$ but also on the other $r$ past states according to the choice of $r$. TAGM can be extended to higher order sequential relationships. We consider a homogeneous Markov process of order $r\in\mathbb{Z}^+$ over a finite state set $\{1,\dots,K\}$ with hidden sequence $\{\+z\}_{n=1}^N$. This stochastic process satisfies 
\[
p(\+z_n|\{\+z_\ell\}_{\ell<n})=p(\+z_n|\{\+z_\ell\}_{\ell=n-r}^{n-1})
\]
or in other words $\+z_n$ can depend on a different number of hidden past states, and we assume that the process is homogeneous i.e., the transition probability 
is independent of $n$. To be as more general as possible we allow that the emission probability of $\+x_n$ can depend not only on $\+z_n$ but also from the previous $m\in\mathbb{Z}^+$ sequence of states
\[
p(\+x_n|\{\+x_\ell\}_{\ell<n},\{\+z_\ell\}_{\ell\leq n})= p(\+x_n|\{\+z_\ell\}_{\ell =n-(m-1)}^n).
\]
Each observation is conditionally independent of the previous ones and of the state sequence history, given the current and the preceding $m-1$ states.

The idea is transform the High order hidden Markov model (HHMM) to a first order hidden Markov model (HMM). It can be done by considering the following two propositions where we omit the prove but it can be found in  \cite{hadar2009high} 
\begin{proposition}
Let $\+Z_n = [\+z_n,\+z_{n-1},\dots,\+z_{n-(\nu-1)}]^\top$. The process $\{\+Z_n\}$ is a first order homogeneous Markov process for any $\nu\geq r$, taking values in $\mathcal{S}^\nu$.
\end{proposition}
\begin{proposition}\label{prop2}
Let $\nu= \max\{ r,m\}$. The state sequence $\{ \+Z_n\}$ and the observation sequence $\{ \+x_n\} $ satisfy
\[
p(\+x_n|\{\+x_\ell\}_{\ell<n},\{\+z_\ell\}_{\ell\leq n})= p(\+x_n|\+Z_n)
\]
and thus constitute a first order  HMM.
\end{proposition}

We can  thus reformulate HHMM as a first order HMM with $K^\nu$ states, where $\nu = \max\{r,m\}$. 
Note that the last $\nu-1$ entries of $\+Z_n$ are equal to the first $\nu-1$ entries of $\+Z_{n+1}$, one concludes that a transition from $\+z_n$ to $\+z_{n+1}$ is possible only if $\lfloor \+z_n/K\rfloor=\+z_{n+1}-\lfloor \+z_{n+1}/K^{\nu-1}\rfloor K^{\nu-1}$, and thus
\[
A_{i,j}=0\quad\text{if }\Big\lfloor \frac{i}{K}\Big\rfloor\not = j-\Big\lfloor \frac{1}{K^{\nu-1}}\Big\rfloor K^{\nu-1}.
\]
Therefore we can use the EM algorithm to find the optimal parameters changing the number of states to $K^\nu$. The $z_{n,j}$s which contribute to the M step for a quantity of state $i$ are given by the set
\[
\mathcal{I}_m(i)=\Bigg\{\Big\lfloor\frac{i}{K^{\nu-m}}\Big\rfloor K^{\nu-m},\Big\lfloor\frac{i}{K^{\nu-m}}\Big\rfloor K^{\nu-m}+1,\dots,\Big(\Big\lfloor\frac{i}{K^{\nu-m}}\Big\rfloor +1\Big)K^{\nu-m}-1\Bigg\}.
\]
Therefore the means become
\[
\mu_i=\frac{\sum_{n=1}^N\+x_n\sum_{j\in\mathcal{I}_m(i)}\gamma(z_{n,j})}{\sum_{n=1}^N\sum_{j\in\mathcal{I}_m(i)}\gamma(z_{n,j})}
\]
The empirical covariances to substitute in the graphical lasso equation is
\[
\Tilde{S}_i= \frac{\sum_{n=1}^N(\+x_{n}-\+\mu_i)(\+x_{n}-\+\mu_i)^\top\sum_{j\in\mathcal{I}_m(i)}\gamma(z_{n,j})}{\sum_{n=1}^N\sum_{j\in\mathcal{I}_m(i)}\gamma(z_{n,j})},
\]
with the hyper-parameter $\tilde{\lambda}_k=\frac{\lambda}{\sum_{n=1}^N\sum_{j\in\mathcal{I}_m(i)}\gamma(z_{n,j})}$. The transition probability matrix becomes
\[
A_{i,j}=
\begin{cases}
&\frac{\sum_{n=2}^N\sum_{k\in\mathcal{I}_r(i)}\xi(z_{n-1,k},z_{n,\lfloor\frac{k}{K}\rfloor+\lfloor\frac{j}{K^{\nu-1}}\rfloor K^{\nu-1}})}{ \sum_{l=1}^{K^\nu}\sum_{n=2}^N\sum_{k\in\mathcal{I}_r(i)}\xi(z_{n-1,k},z_{n,l})}\quad\text{if }\lfloor\frac{i}{K}\rfloor=j-\lfloor\frac{j}{K^{\nu-1}}\rfloor K^{\nu-1},\\
&0\quad\text{otherwise}.
\end{cases}
\]
Finally,  the initial state probabilities are given by
\[
\pi_i=\gamma(z_{1,i}).
\]
Therefore if we increase the number of states to $K^{\nu}$ and modify the TAGM M step formulas with the one just found we obtain the MemTAGM. We test MemTAGM performance in subsection 9.2 where we compare it with TAGM.

\subsection{On-line learning: Incremental Time Adaptive Gaussian model (IncTAGM)}
In many applications it is important to update the TAGM parameters almost instantaneously every time a new observation comes up. Since TAGM does not allow to have such a quick response, it can be extended to an incremental version. We call this model IncTAGM and it initially starts as a standard TAGM reading a set of observations and updating its current parameters $\pi,A, \mu, \+\Theta$ according to new incoming data. Therefore, after the standard TAGM has finished training on its observation set, it calculates the revised $\alpha, \beta, \xi$ and $\gamma$ variables based on the new set of observations. To update the model incrementally, we need recursive equations for $\alpha$ and $\beta$ which depend on past values. Notice that the $\alpha$ recursive equation is already of this form. While the $\beta$ recursive equation needs an approximation to become of that form. In fact if we assume that $\beta(z_{T,i})\simeq \beta(z_{T,j})$ for every $i\not=j$, the $\beta$ recursive equation becomes
\begin{equation}
\beta(\+z_{T+1}) =\frac{\beta(\+z_{T}) }{\sum_{\+z_{T+1}}p(\+x_{T+1}|\+z_{T+1})p(\+z_{T+1}|\+z_T)}.
\end{equation}

The M step optimal parameters are updated in the following way:

the initial state $\pi$
\begin{equation}
\pi_k' =\gamma(z_{1,k}),
\end{equation}
the transition matrix $\+A$ 
\begin{align}
A^{T+1}_{j,k} &= \frac{ \sum_{n=2}^T\xi(z_{n-1,j},z_{n,k})+\xi(z_{T,j},z_{T+1,k})}{\sum_{l=1}^K\sum_{n=2}^T\xi(z_{n-1,j},z_{n,l})+\sum_{l=1}^K\xi(z_{T,j},z_{T+1,l})}\notag\\
&= \frac{ \sum_{n=2}^T\xi(z_{n-1,j},z_{n,k})}{\sum_{n=2}^{T+1}\gamma(z_{n-1,j})} + \frac{\xi(z_{T,j},z_{T+1,k})}{\sum_{n=2}^{T+1}\gamma(z_{n-1,j})}\notag\\
&= \frac{ \sum_{n=2}^{T}\gamma(z_{n-1,j})}{\sum_{n=2}^{T+1}\gamma(z_{n-1,j})} \frac{ \sum_{n=2}^T\xi(z_{n-1,j},z_{n,k})}{\sum_{n=2}^{T}\gamma(z_{n-1,j})}  + \frac{\xi(z_{T,j},z_{T+1,k})}{\sum_{n=2}^{T+1}\gamma(z_{n-1,j})}\notag\\
&= \frac{ \sum_{n=2}^{T}\gamma(z_{n-1,j})}{\sum_{n=2}^{T+1}\gamma(z_{n-1,j})}A^{T}_{j,k} + \frac{\xi(z_{T,j},z_{T+1,k})}{\sum_{n=2}^{T+1}\gamma(z_{n-1,j})}.\label{recA}
\end{align}
Note that if we sum by $k\in\{1,\dots,K\}$ the row normalization holds. Similarly we obtain the formula for the means
\begin{equation}
\mu_k^{T+1}=\frac{ \sum_{n=1}^{T}\gamma(z_{n,k})}{\sum_{n=1}^{T+1}\gamma(z_{n,k})}\mu_k^T+\frac{\gamma(z_{T+1,k})\+x_{T+1}}{\sum_{n=1}^{T+1}\gamma(z_{n,k})}
\end{equation}
and the empirical covariances
\begin{equation}
\Tilde{S}^{T+1}_k=\frac{ \sum_{n=1}^{T}\gamma(z_{n,k})}{\sum_{n=1}^{T+1}\gamma(z_{n,k})}\Tilde{S}^{T}_k+ \frac{\gamma(z_{T+1,k})(\+x_{T+1}-\+\mu^{T+1})(\+x_{T+1}-\+\mu^{T+1})^\top}{\sum_{n=1}^{T+1}\gamma(z_{n,k})},
\end{equation}
 with the hyper-parameter $\tilde{\lambda}_k=\frac{\lambda}{\sum_{n=1}^{T+1}\gamma(z_{n,k})}$.

\subsubsection{Slide Incremental Time Adaptive Gaussian model (S-IncTAGM)}
The training of new data points results in the accumulation of an increasingly large observation set. As a result, if the time sequence considered is large and the first point is far away in the past respect to the last point it is possible that the initial trained observation points become outdated after many updates and therefore they do not carry any useful information to analyze the more recent points.  Therefore, the new addition to the IncTAGM is a fixed sliding window to effectively analyze discrete data (appropriately discarding the outdated observations) whilst updating its model parameters. 

The estimation of the $\alpha, \beta, \xi$ and $\gamma$ variables remains the same as in the IncTAGM algorithm. What changes are the $A$, $\+\mu$ and $\+\Theta$ updates. Using the \textit{simple moving average} (SMA) definition
\begin{equation}
sma=\frac{x_1+x_2\dots+x_n+x_{n+1}-x_1}{n}= ave +\frac{x_{n+1}}{n}-\frac{x_1}{n}
\end{equation}
where $ave= \frac{x_1+x_2\dots+x_n}{n}$,  we update $A$, $\+\mu$ and $\+\Theta$ in the following way
\begin{align}
A^{T+1}_{j,k} &= \frac{\sum_{n=3}^{T+1}\xi(z_{n-1,j},z_{n,k})}{\sum_{n=3}^{T+1}\gamma(z_{n-1,j})},\\
\mu_k^{T+1} &=\frac{ \sum_{n=2}^{T+1}\gamma(z_{n,k})\+x_n}{\sum_{n=2}^{T+1}\gamma(z_{n,k})},\\
\Tilde{S}^{T+1}_k&=\frac{ \sum_{n=2}^{T+1}\gamma(z_{n,k})(\+x_{n}-\+\mu^{T+1})(\+x_{n}-\+\mu^{T+1})^\top}{\sum_{n=2}^{T+1}\gamma(z_{n,k})}
\end{align}
with $\tilde{\lambda}_k=\frac{\lambda}{\sum_{n=2}^{T+1}\gamma(z_{n,k})}$.




\section{Conclusions}
We presented a novel methodology to perform data-mining and forecasting on multi-variate time-series. Our method, namely TAGM, combines HMMs and GGMs, providing a way to simultaneously cluster non-stationary time-series into stationary sub-groups and for each cluster detecting probability relationships among variables through graphical model inference. This simultaneous inference is suitable to be transformed into a time-varying regression model that allows to make predictions on non-stationary time-series.
The coupling we performed allows to generalize many state-of-the-art methods and provide a wide range of analysis type to be performed on the time series. 

\bibliographystyle{plain}
\bibliography{bibliography}

\end{document}